\documentclass{article}

\usepackage[final]{cpal_2026}

\usepackage{url}
\usepackage{authblk}
\usepackage{amssymb}
\usepackage{amsmath}
\usepackage{mathtools}
\usepackage{lineno}
\usepackage[ruled,vlined]{algorithm2e}
\usepackage{booktabs}
\usepackage[hidelinks]{hyperref}
\usepackage{caption}
\usepackage{subcaption}
\usepackage{graphicx}
\usepackage{tikz}
\usepackage{wrapfig}

\title{Simplex Deep Linear Discriminant Analysis}

\author[1,4]{Arman Bolatov}
\author[2]{Maxat Tezekbayev}
\author[3,4]{Zhenisbek Assylbekov}

\affil[1]{Mohamed bin Zayed University of Artificial Intelligence}
\affil[2]{Nazarbayev University}
\affil[3]{Purdue University Fort Wayne}
\affil[4]{National Center of Space Research and Technology, Almaty, Kazakhstan}
\affil[ ]{\texttt{arman.bolatov@mbzuai.ac.ae, maxat.tezekbayev@alumni.nu.edu.kz, zassylbe@pfw.edu}}

\begin{document}

\maketitle

\begin{abstract}
We revisit Deep Linear Discriminant Analysis (Deep LDA) from a likelihood-based perspective. While classical LDA is a simple Gaussian model with linear decision
boundaries, attaching an LDA head to a neural encoder raises the question of how to train the resulting deep classifier by maximum likelihood estimation (MLE). We first show that end-to-end MLE training of an unconstrained Deep LDA model ignores discrimination: when both the LDA parameters and the encoder parameters are learned jointly, the likelihood admits a degenerate solution in which some of the class clusters may heavily overlap or even collapse, and classification performance deteriorates. Batchwise moment re-estimation of the LDA parameters does not remove this failure mode. We then propose a constrained Deep LDA formulation that fixes the class means to the vertices of a regular simplex in the latent space and restricts the shared covariance to be spherical, leaving only the priors and a single variance parameter to be learned along with the encoder. Under these geometric constraints, MLE becomes stable and yields well-separated class clusters in the latent space. On images (Fashion-MNIST, CIFAR-10, CIFAR-100) and texts (AG News, CLINC150), the resulting Deep LDA models achieve accuracy competitive with softmax baselines while offering a simple, interpretable latent geometry that is clearly visible in two-dimensional projections.
\end{abstract}

\section{Introduction}\label{sec:intro}

Linear Discriminant Analysis (LDA) traces its roots to the classical works of \citet{fisher1936use} and \citet{rao1948}. It remains one of the most widely used classification methods and is valued for its interpretability. LDA assumes that the class label \(Y \in \{1,\dots,C\}\) follows a categorical prior with probabilities \(\pi = (\pi_1,\dots,\pi_C)\) on the probability simplex:
\begin{equation}
\Pr(Y=c)=\pi_c, 
\qquad 
\pi_c\ge0,\ \ \sum_{c=1}^C \pi_c = 1 .
\label{eq:priors}
\end{equation}
Conditional on the class, the feature vector \(X \in \mathbb{R}^d\) is assumed Gaussian with a \emph{shared} covariance matrix \(\Sigma \in \mathbb{R}^{d\times d}\):
\begin{equation}
X \mid Y=c \;\sim\; \mathcal{N}(\mu_c,\;\Sigma), \qquad c=1,\dots,C,
\label{eq:class_cond}
\end{equation}
so that each class has its own mean \(\mu_c \in \mathbb{R}^d\) while the covariance \(\Sigma\) is common to all classes.

\begin{figure}[htbp]
  \begin{minipage}[t]{.47\textwidth}
  \centering
  \includegraphics[width=\textwidth]{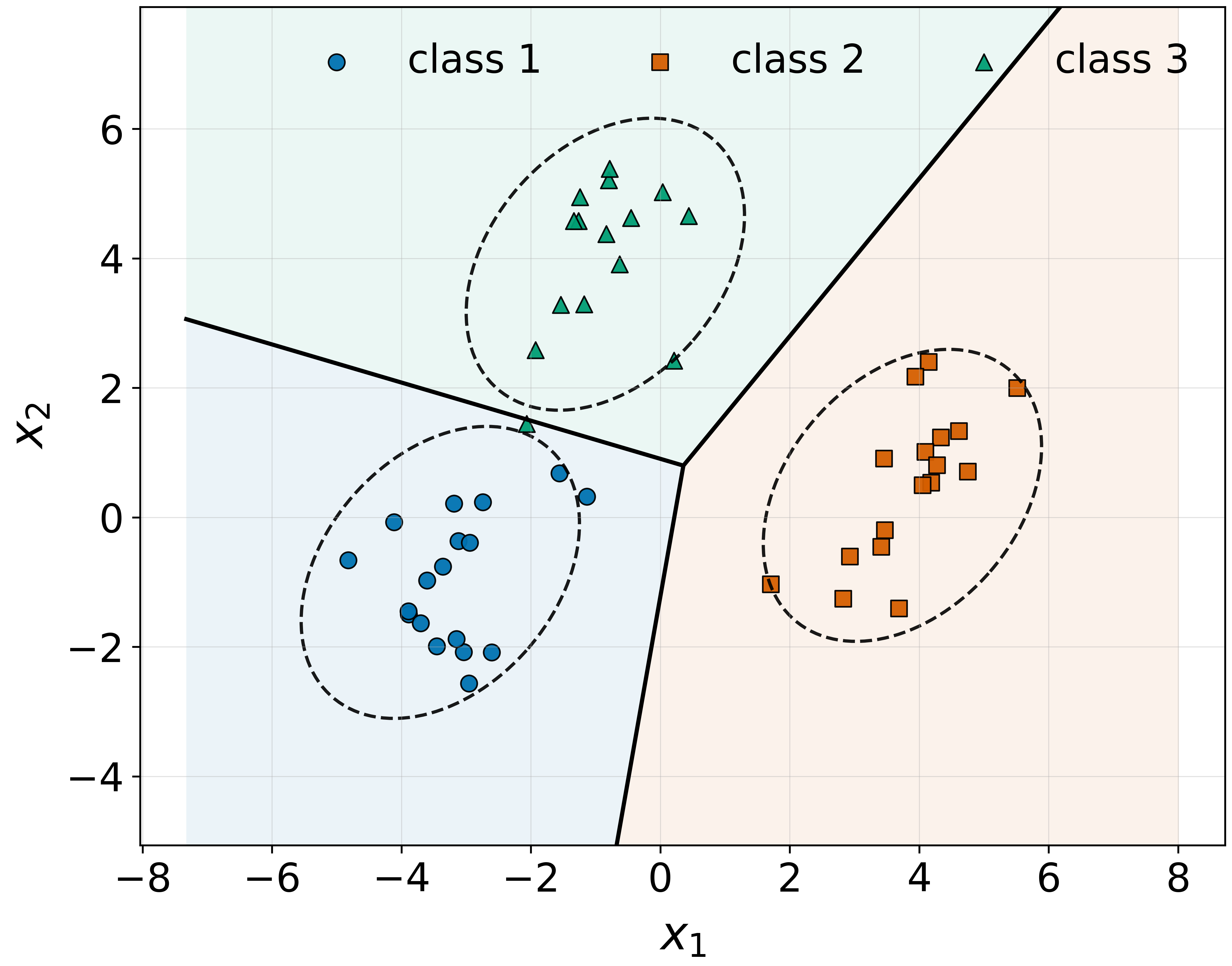}
  \caption{Three-class LDA example with shared covariance. Ellipses that contain 90\% of the probability for each of the three classes are shown. Faint colors show decision regions, and solid lines mark decision boundaries.}
  \label{fig:lda-sample}
  \end{minipage}\hfill\begin{minipage}[t]{.47\textwidth}
  \centering
  \includegraphics[width=\textwidth]{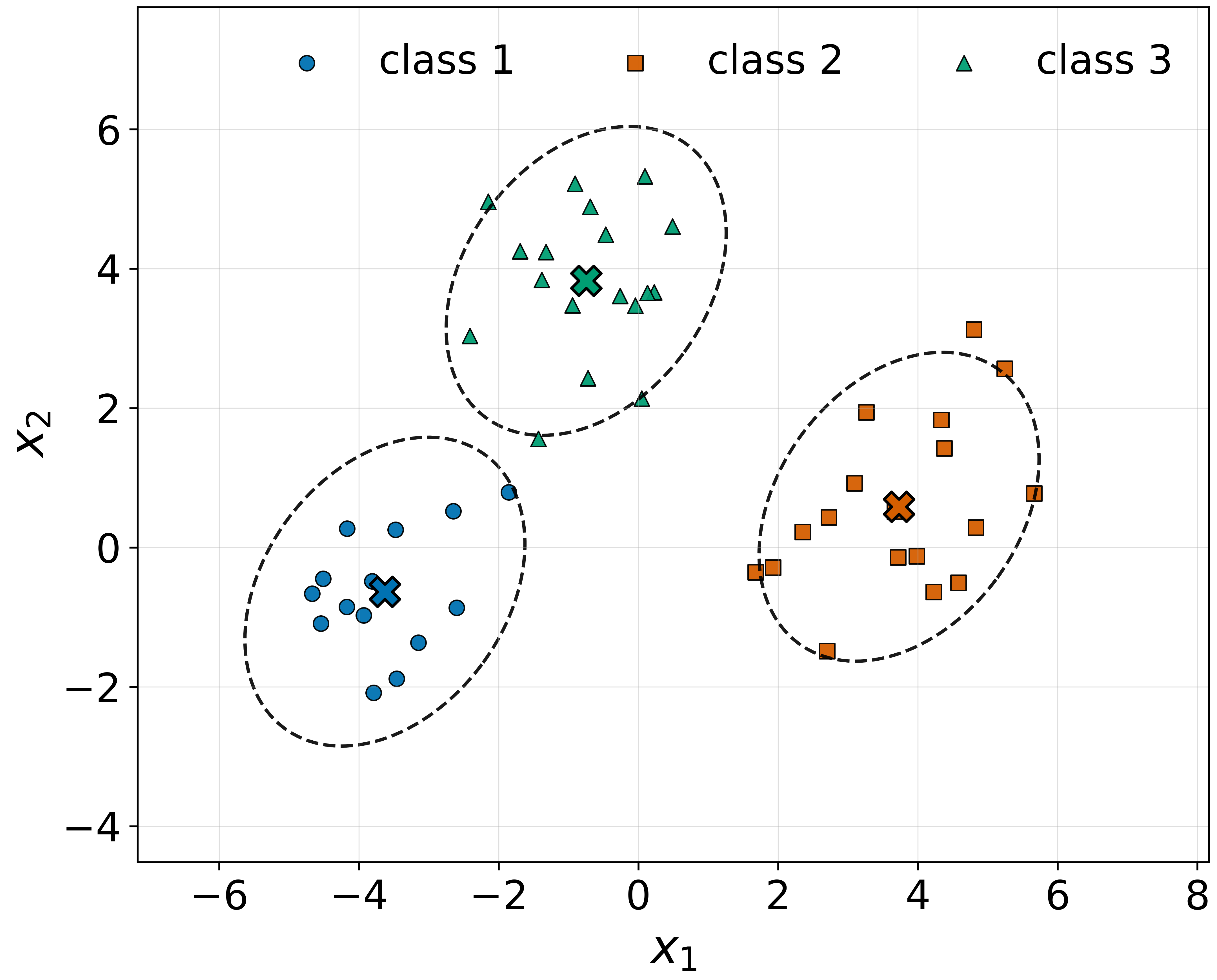}
  \caption{Gradient-based ML fit of a three-class LDA model ($n=50$). Dashed ellipses show the 90\% contours of the learned Gaussian classes, and filled ``X'' markers indicate the estimated class centers. The recovered parameters nearly coincide with the closed-form MLE.}
  \label{fig:lda_result}    
  \end{minipage}
\end{figure}

By specifying the marginal law of \(Y\) and the class-conditional law of \(X\mid Y\), LDA defines a joint probabilistic model over \((X,Y)\). As a generative model, its parameters \(\big(\pi,\{\mu_c\}_{c=1}^C,\Sigma\big)\) can be estimated by maximum likelihood. Once fitted, prediction for a new sample \(x\) proceeds via the posterior class probabilities
\(\Pr(Y=c \mid X=x)\), obtained from the prior \(\Pr(Y=c)\) and the class-conditionals \(p(x\mid Y=c)\) by Bayes' rule.

Because LDA posits an explicit data-generating story, it offers clear interpretability. A well-known limitation, however, is that the resulting decision boundaries are \emph{linear}  between any pair of classes (Figure~\ref{fig:lda-sample}), which can be too restrictive when classes are not linearly separable in the input space.

To address this, \emph{Deep LDA} \citep{DBLP:journals/corr/DorferKW15} couples a learnable representation with an LDA head. Concretely, one first maps inputs into a latent space via a neural encoder \(Z=f_\psi(X)\) parameterized by \(\psi\). In that latent space, the pair \((Z,Y)\) is modeled according to the LDA assumptions in \eqref{eq:priors}--\eqref{eq:class_cond} with class means \(\{\mu_c\}\) and a shared covariance \(\Sigma\). In effect, Deep LDA replaces the conventional softmax layer with an LDA head, while letting the encoder learn a representation in which the LDA model is appropriate.

In contrast to a maximum-likelihood treatment, Deep LDA~\citep{DBLP:journals/corr/DorferKW15} optimizes
Fisher's generalized eigenvalue objective in the latent space. A straightforward maximization, however, turned out to be unstable: as noted by \citet{DBLP:journals/corr/DorferKW15}, the
network tends to inflate a single largest eigenvalue and essentially ignore poorly separated
classes, which leads to trivial solutions. To mitigate this, they introduce a
heuristic variant of the criterion that focuses only on the eigen-directions with small
discriminative power (see their Eq.~(9)), departing from the classical Fisher objective.

However, LDA is a classical parametric probabilistic model, and such models are typically trained with maximum likelihood estimation (MLE). This naturally raises the question:

\begin{quote}
\emph{Why not train Deep LDA by maximum likelihood, using standard gradient-based methods end-to-end?}
\end{quote}

However, as we show in Section~\ref{sec:mle}, when both the class means and the shared covariance are learned jointly with a flexible encoder, the likelihood admits a maximizer that ignores discrimination: the encoder not only collapses each sample onto its corresponding class mean, but also may drive some of the class means  toward each other so that the resulting clusters become arbitrarily tight and heavily overlapping. This yields an arbitrarily high training likelihood but destroys the geometric structure required for successful discrimination.

The failure is a matter of structural degeneracy of the model. Even replacing gradient updates of the LDA parameters with batchwise moment estimates does not prevent collapse: the encoder can always shrink between-class variation, and nothing in the free LDA parameterization opposes this.

These observations motivate a different standpoint. Instead of allowing the LDA head to adapt freely to the encoder, we constrain its geometry so that maximum likelihood becomes well-behaved in the deep setting. In Section~\ref{sec:success-deep-lda} we introduce a simple and effective modification: fix the class means to the vertices of a regular simplex and restrict the shared covariance to be spherical. Under these constraints, the likelihood no longer admits degenerate solutions, and end-to-end MLE becomes stable. The encoder learns to map inputs to well-separated clusters around fixed prototypes, preserving the probabilistic structure of LDA while avoiding the collapse seen in the unconstrained model.

The remainder of the paper develops this perspective and demonstrates its practical viability on both synthetic and real datasets. Our code is available at \url{https://github.com/zh3nis/simplex-lda}.

\section{Maximum-Likelihood Training: A First Look}\label{sec:mle}
As a warm-up, we first verify that the classical (non-deep) LDA admits accurate MLE via gradient-based optimization on synthetic data.

\subsection{Warm-up on Classical LDA}
We consider the data generation  model in \eqref{eq:priors} and \eqref{eq:class_cond}. 
Let \(\theta \coloneqq \big(\pi,\{\mu_c\}_{c=1}^C,\Sigma\big)\) denote the parameters with \(\pi\) on the probability simplex and \(\Sigma\succ 0\). Given a sample
\(\{(x_1,y_1),\dots,(x_n,y_n)\}\) drawn from \eqref{eq:priors}--\eqref{eq:class_cond}, the {log-likelihood} is
\begin{equation}
\label{eq:avg-loglike}
\mathcal{L}(\theta)
=\frac{1}{n}\sum_{i=1}^n \log\!\Big( \pi_{y_i}\,\phi(x_i;\,\mu_{y_i},\Sigma) \Big),
\end{equation}
where, for \(x\in\mathbb{R}^d\),
\(
\phi(x;\mu,\Sigma)
=(2\pi)^{-d/2}\,(\det\Sigma)^{-1/2}
\exp\left\{-\tfrac{1}{2}(x-\mu)^\top\Sigma^{-1}(x-\mu)\right\}
\)
is the \(\mathcal{N}(\mu,\Sigma)\) density. 

\paragraph{Optimization setup.}
We maximize the log-likelihood \(\mathcal{L}(\theta)\) with Adam, treating its negative $-\mathcal{L}(\theta)$ as a loss. In practice it is convenient to use unconstrained reparameterizations:
(i) logits \(\alpha\in\mathbb{R}^C\) with \(\pi=\mathrm{softmax}(\alpha)\), and
(ii) a Cholesky factorization \(\Sigma=LL^\top\) with \(L\) lower triangular and positive diagonal. This enforces \(\pi\in\Delta^{C-1}\) and \(\Sigma\succ 0\) while enabling standard auto-differentiation.

\paragraph{Result.}
On synthetic data generated from \eqref{eq:priors}--\eqref{eq:class_cond}, gradient descent with Adam reliably converges to the maximum-likelihood solution; see Figure~\ref{fig:lda_result}. This confirms that, at least for the classical LDA model, direct gradient-based likelihood maximization is effective and accurate.

\paragraph{Relation to Fisher's criterion.}
Classical LDA can also be derived by maximizing Fisher's generalized eigenvalue objective
\begin{equation}
  \label{eq:fisher-criterion}
  \max_{W \in \mathbb{R}^{d \times (C-1)}} 
  J(W)
  \coloneqq 
  \operatorname{tr}\!\big( (W^\top S_W W)^{-1} W^\top S_B W \big),
\end{equation}
where $S_W$ and $S_B$ are the within-class and between-class scatter matrices.
For Gaussian class-conditional models with shared covariance, maximizing
\eqref{eq:fisher-criterion} yields the same discriminant subspace---and
hence the same decision boundaries---as the maximum-likelihood solution of
the generative LDA model.

To verify this equivalence in our setting, we repeated the 
experiment above but trained the encoder by optimizing
Fisher's criterion instead of the log-likelihood.
As expected, the learned decision boundaries, and recovered
parameters matched those obtained from maximum-likelihood training,
confirming that both objectives recover the same classifier in the
shared-covariance Gaussian regime.

\subsection{A Cautionary Experiment: Likelihood Training of Deep LDA}

For the Deep LDA experiment, we again generate a dataset 
\(\{(x_1,y_1),\ldots,(x_n,y_n)\}\) 
from the model in \eqref{eq:priors}--\eqref{eq:class_cond}. The encoder \(f_\psi(\cdot)\) is a two-layer feed-forward network with ReLU activations and 32 hidden units. The resulting log-likelihood is
\begin{equation}
\label{eq:deep-lda-like}
\mathcal{L}(\theta)
=\frac{1}{n}\sum_{i=1}^n
\log\!\left(
\pi_{y_i}\,
\phi\!\left(f_\psi(x_i);\ \mu_{y_i},\,\Sigma\right)
\right),
\end{equation}
where \(\theta := (\psi, \pi, \{\mu_c\}_{c=1}^C, \Sigma)\).

\begin{figure}[htbp]
  \begin{minipage}[t]{.47\textwidth}
  \centering
  \includegraphics[width=\textwidth]{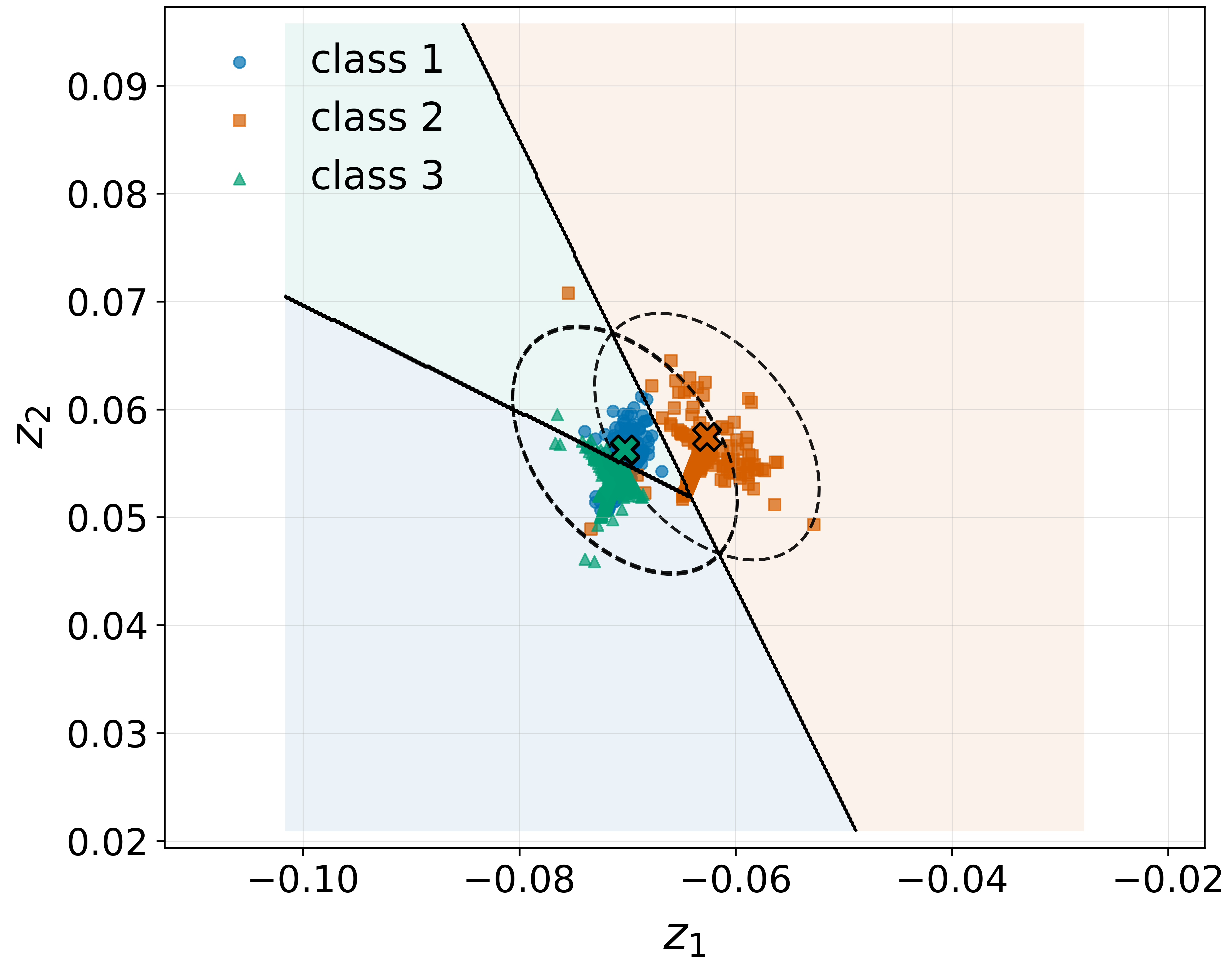}
  \caption{Deep LDA embeddings \(z_i=f_\psi(x_i)\) after likelihood training. 
  Two classes collapse; samples lie near their class centroids \(\{\mu_c\}\) with a nearly singular shared covariance \(\Sigma\) ($|\Sigma|\approx3\cdot10^{-10}$). 
  Training accuracy: 67.1\%; test accuracy: 66.6\%.}
  \label{fig:deep_lda_emb}
  \end{minipage}
  \hfill
  \begin{minipage}[t]{.47\textwidth}
  \centering
  \includegraphics[width=\linewidth]{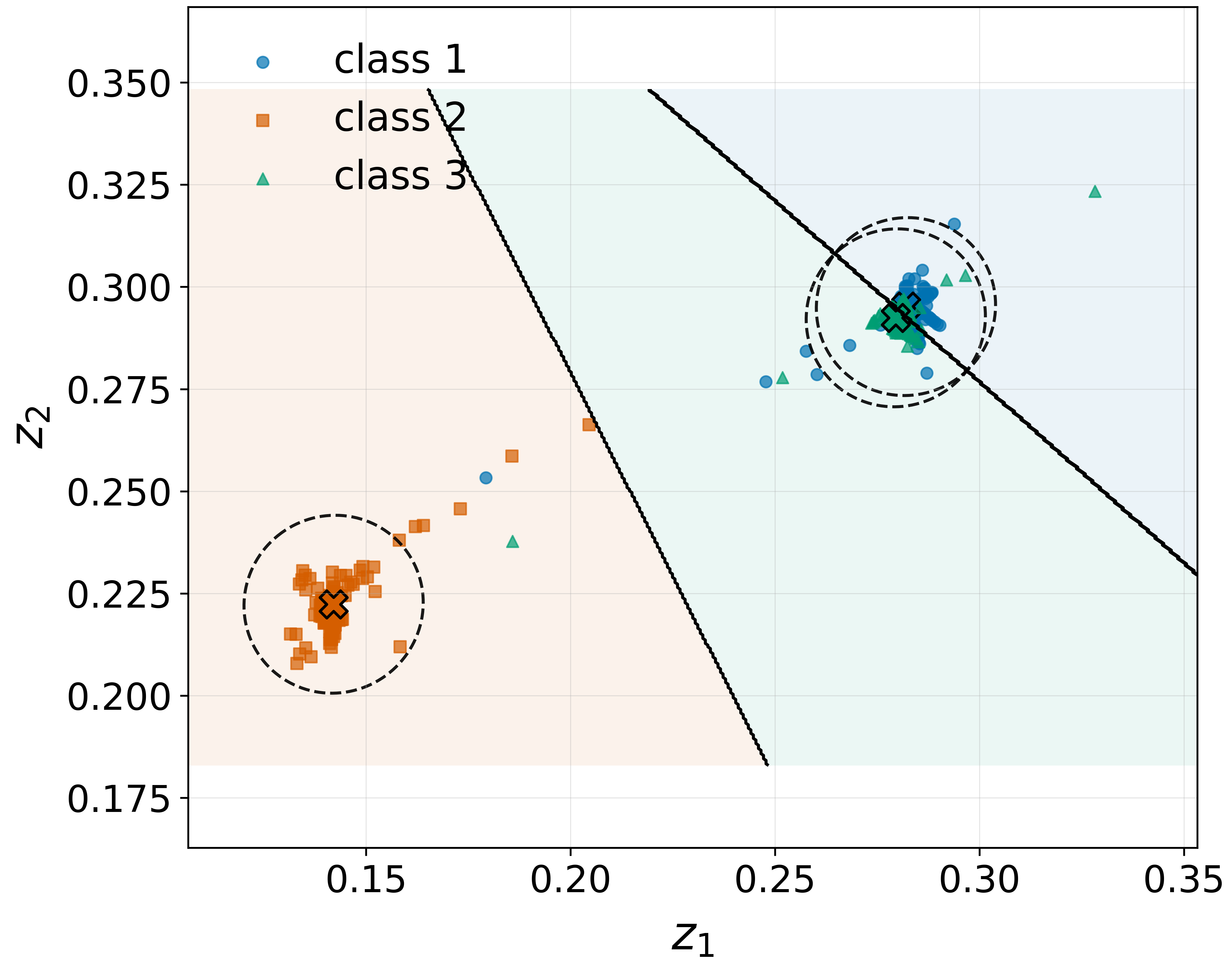}
  \caption{Deep LDA embeddings \(f_\psi(x_i)\) after likelihood training with batchwise LDA re-estimation. 
  Two class clusters collapse; embeddings concentrate near their centroids \(\{\mu_c\}\) while the shared covariance becomes ill-conditioned ($|\Sigma|\approx10^{-8}$). 
  Training accuracy: 67.8\%; test accuracy: 66.6\%.}
  \label{fig:deep_lda_emb2}    
  \end{minipage}
\end{figure}

We generate \(20{,}000\) training and \(4{,}000\) test samples. 
Training is run for 100 epochs using Adam with PyTorch's default hyperparameters and a minibatch size of 256 (batch size 1024 at evaluation). The learned embeddings \(f_\psi(x_i)\) are shown in Figure~\ref{fig:deep_lda_emb}.

The pathology is immediate: the optimizer does not enforce separation among all classes—two of the class clusters collapse into one. Moreover, for each labeled pair \((x, y)\), the encoder maps \(x\) extremely close to its class centroid \(\mu_{y}\), driving the squared Mahalanobis distance
\[
\left(f_\psi(x)-\mu_{y}\right)^\top
\Sigma^{-1}
\left(f_\psi(x)-\mu_{y}\right)
\]
toward zero, while \(\Sigma\) simultaneously shrinks toward near-singularity.
Both effects inflate the Gaussian term 
\(\phi(f_\psi(x);\mu_{y},\Sigma)\), yielding a high training likelihood.\footnote{We note that replacing the log-likelihood with Fisher's criterion does not
alleviate this issue: in our experiments, optimizing the Fisher objective
produced the same collapse of embeddings and class means as the
maximum-likelihood formulation. A similar observation was made by \citet{DBLP:journals/corr/DorferKW15}.}

This behavior exposes a fundamental mismatch between the maximum-likelihood objective and the goal of accurate classification: the training and test accuracies are only 67.1\% and 66.6\%, respectively.

\subsection{Batchwise LDA Re-estimation}\label{sec:ema}
In an attempt to mitigate the collapse observed under end-to-end likelihood training, we stopped updating the LDA-head parameters
\(\bigl(\pi, \{\mu_c\}_{c=1}^C,\Sigma\bigr)\) by gradient descent and instead \emph{estimated} them on mini-batches, while training
only the encoder \(f_\psi\). Let \(\{(x_1,y_1),\ldots,(x_m,y_m)\}\) be the current batch and set \(z_i=f_\psi(x_i)\).
Define the batch index sets \(I_c=\{i\in\{1,\dots,m\}:y_i=c\}\) and \(n_c=|I_c|\).
The plug-in LDA estimates on this batch are
\begin{align*}
\widehat{\pi}_c^{(t)}
&=\frac{n_c}{m},\qquad \widehat{\mu}_c^{(t)}
=\frac{1}{n_c}\sum_{i\in I_c} z_i,\qquad c=1,\dots,C,\\ 
\widehat{\Sigma}^{(t)}
&=\frac{1}{m}\sum_{c=1}^C \sum_{i\in I_c}
\bigl(z_i-\widehat{\mu}_c^{(t)}\bigr)\bigl(z_i-\widehat{\mu}_c^{(t)}\bigr)^\top.
\end{align*}

Because successive batches can vary substantially, we smooth these estimates with an exponential moving average (EMA).
Writing tildes for the smoothed parameters, we use
\begin{align*}
\widetilde{\pi}_c^{(t)}
&=\beta\,\widetilde{\pi}_c^{(t-1)} + (1-\beta)\,\widehat{\pi}_c^{(t)}, 
\qquad c=1,\dots,C,\\
\widetilde{\mu}_c^{(t)}
&=\beta\,\widetilde{\mu}_c^{(t-1)} + (1-\beta)\,\widehat{\mu}_c^{(t)}, 
\qquad c=1,\dots,C,\\
\widetilde{\Sigma}^{(t)}
&=\beta\,\widetilde{\Sigma}^{(t-1)} + (1-\beta)\,\widehat{\Sigma}^{(t)},
\end{align*}
with \(\beta\in[0,1)\) (e.g., \(\beta=0.9\)). At iteration \(t\), the likelihood in \eqref{eq:deep-lda-like} retains the same form,
but now only the encoder parameters are optimized, i.e., \(\theta:=\psi\), while
\(\{\pi_c,\mu_c\}_{c=1}^C,\Sigma\) are the EMA-smoothed re-estimates
\(\{\widetilde{\pi}_c^{(t)},\widetilde{\mu}_c^{(t)}\}_{c=1}^C,\widetilde{\Sigma}^{(t)}\).

\paragraph{Result.}
We run this modified Deep LDA training procedure on data generated from \eqref{eq:priors}--\eqref{eq:class_cond}.  
Unfortunately, the batchwise re-estimation with EMA does not resolve the collapse; see Figure~\ref{fig:deep_lda_emb2}.  
Two of the three classes still collapse.  
For each labeled pair \((x,y)\), the encoder continues to map  
\(x\) to an embedding \(z=f_\psi(x)\) extremely close to its class mean \(\mu_y\),  
while the pooled covariance \(\Sigma\) contracts toward near-singularity  
(\(|\Sigma|\approx 10^{-8}\)).  
Both effects again inflate the Gaussian likelihood term, resulting in a large training objective but poor classification performance.\footnote{A similar experiment using Fisher's criterion in place of the log-likelihood
showed no improvement: the same degeneracy persisted despite the batchwise
moment updates.}

We hypothesize that the issue is not with gradient-based MLE itself---which we still wish to retain---but with the unconstrained Deep LDA parameterization.
This leads to the central question: \emph{how should Deep LDA be constrained so that MLE learns an accurate classifier?}

\section{Successful Training of Deep LDA with MLE}
\label{sec:success-deep-lda}

The failures observed in Section~\ref{sec:mle} stem from the excessive flexibility of the deep model: all the LDA parameters ($\pi$, ${\mu_c}$ and  $\Sigma$) are free to adapt to the encoder $f_\psi$, allowing not only the embeddings to collapse tightly around their class centroids but also the class centroids themselves to drift closer together, producing overlapping clusters. In this section we show that likelihood-based training can lead to discrimination by \emph{constraining} the geometry of the LDA head.

\paragraph{Simplex-Constrained Means and Spherical Covariance.}

\begin{figure}[htbp]
\begin{minipage}[t]{.47\textwidth}
\centering
\begin{tikzpicture}[scale=2.2]

\coordinate (mu1) at (0,0);         
\coordinate (mu2) at (2,-0.18);     
\coordinate (mu3) at (0.75,0.65);   

\coordinate (mu4) at (1,1.72);

\draw[thick] (mu1) -- (mu2) -- (mu4) -- (mu1);

\draw[thick,dashed] (mu1) -- (mu3) -- (mu2);
\draw[thick,dashed] (mu3) -- (mu4);

\foreach \p in {mu1,mu2,mu3,mu4}
  \fill (\p) circle (1.3pt);

\node[below left]  at (mu1) {$\mu_1$};
\node[below right] at (mu2) {$\mu_2$};
\node[left]        at (mu3) {$\mu_3$};
\node[above]       at (mu4) {$\mu_4$};

\end{tikzpicture}
\caption{Regular simplex in $\mathbb{R}^3$ for $C=4$.
The four fixed class means ${\mu_c}$ form a tetrahedron with edges of equal length; dashed lines indicate hidden edges in the 2D projection.}
\label{fig:simplex_example}
\end{minipage}\hfill\begin{minipage}[t]{.47\textwidth}
  \centering
  \includegraphics[width=.9\linewidth]{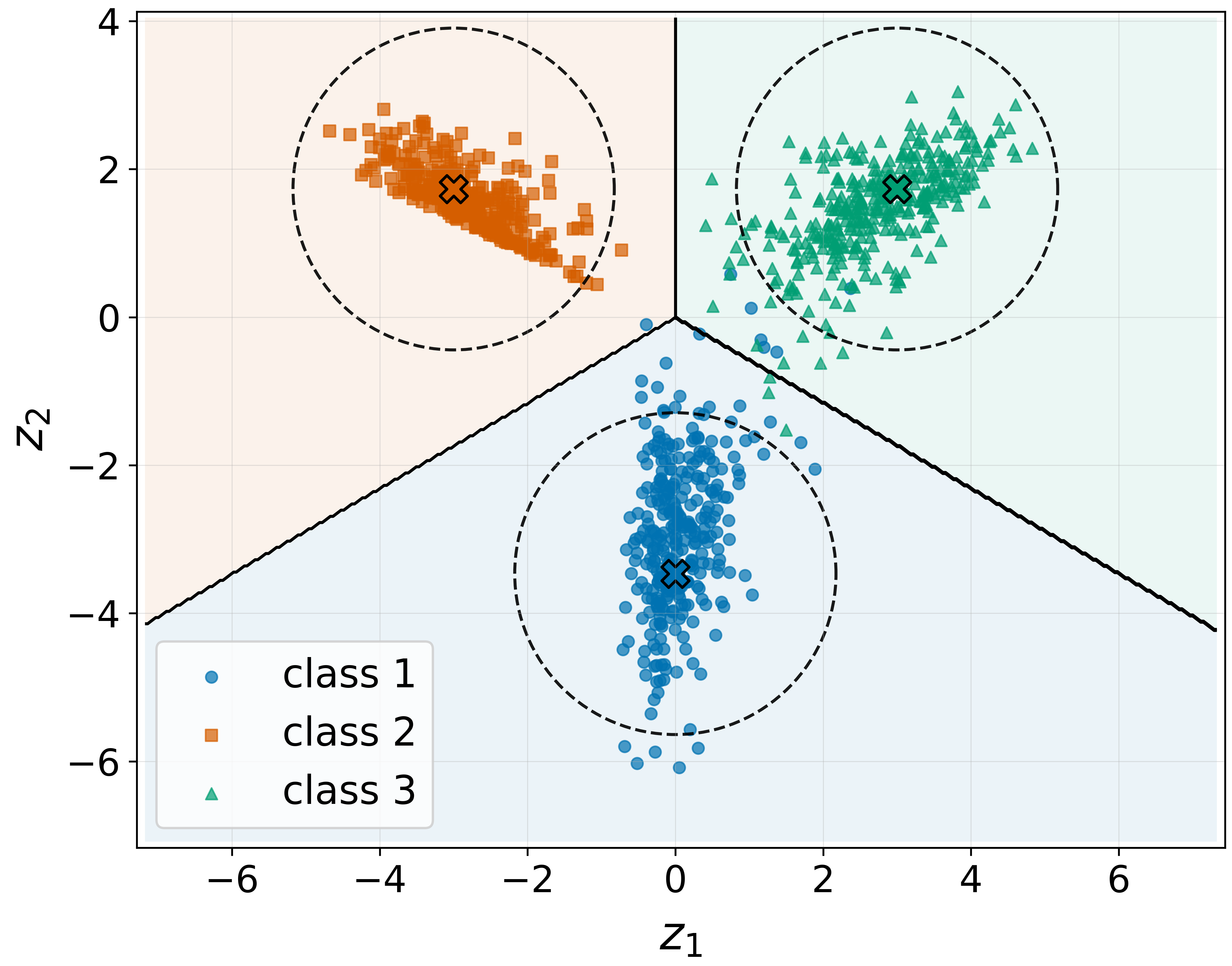}
  \caption{Embeddings $f_\psi(x)$ after training Deep LDA with simplex-constrained means and spherical covariance via MLE. Class-conditional clouds are  well separated around the fixed simplex vertices (marked by crosses). Training set accuracy is 99.2\%, test set accuracy is 99.4\%.}
  \label{fig:deep_lda_emb3}    
\end{minipage}
\end{figure}

We keep the encoder $Z=f_\psi(X)$ as before, but replace the fully free LDA head with a structured variant:

\begin{itemize}
  \item The class means $\{\mu_c\}_{c=1}^C$ are fixed to the vertices of a regular simplex in $\mathbb{R}^{C-1}$, centered at the origin (an example for $C=4$ is given in Figure~\ref{fig:simplex_example}).
  \item The shared covariance is restricted to be spherical: $\Sigma = \sigma^2 I_d$ with a \emph{single} trainable variance parameter $\sigma^2$.
  \item The class priors $\{\pi_c\}_{c=1}^C$ remain learnable, parameterized via unconstrained logits.
\end{itemize}

Concretely, for $C$ classes we first construct a regular simplex in $\mathbb{R}^{C-1}$ whose vertices have unit norm and zero mean. We rescale the simplex so that the Euclidean distance between any pair of class means is a fixed constant $s>0$ (we use $s=6$ so that the classes are well-separated at initialization when we set $\sigma=1$).

The output of the encoder is made $(C-1)$-dimensional, since for $C$ classes LDA needs at most $d=C-1$ dimensions. For an embedding $Z=f_\psi(X)$, the class-conditional model is
\[
Z\mid Y=c \sim \mathcal{N}(\mu_c,\;\sigma^2 I_d),
\qquad c=1,\dots,C.
\]
The squared Mahalanobis distance reduces to a scaled squared Euclidean distance
\[
\|z-\mu_c\|_{\Sigma^{-1}}^2
\;=\;
(z-\mu_c)^\top(\sigma^{-2} I_d)(z-\mu_c)
\;=\;
\frac{1}{\sigma^2}\,\|z-\mu_c\|^2.
\]

A closely related geometric structure has been observed empirically in
deep classifiers trained with cross-entropy. In particular,
\citet{papyan2020prevalence} document the \emph{neural collapse}
phenomenon: in the terminal phase of training, the last-layer features
collapse to their class means, these means themselves converge to the
vertices of a simplex equiangular tight frame, and the classifier
weights align with those means so that prediction effectively reduces
to a nearest-class-center rule. Our simplex-constrained Deep LDA head can be viewed as an
explicit generative realization of a similar latent geometry: the
simplex structure is imposed \emph{a priori} through the
class-conditional model, and the encoder is trained by maximum
likelihood to populate these fixed prototypes.

\paragraph{Discriminant Functions and Likelihood.}

For our simplex-based LDA head, we can write the unnormalized log density of $(Y,Z)$ as
\begin{equation}
\label{eq:delta-simplex}
\delta_c(z)
\;\coloneqq\;
\log \pi_c
-\frac{1}{2}
\left(
\frac{\|z-\mu_c\|_2^2}{\sigma^2}
+d\log\sigma^2
\right),
\qquad c=1,\dots,C.
\end{equation}
Up to an additive constant that does not depend on $c$ or on the parameters, $\delta_c(z)$ equals
\[
\log \Pr(Y=c) + \log \phi(z;\mu_c,\sigma^2 I_d).
\]

Thus, for labeled data $\{(x_i,y_i)\}_{i=1}^n$ and embeddings $z_i=f_\psi(x_i)$, the log-likelihood 
is (up to an additive constant)
\begin{equation}
\label{eq:deep-lda-mle}
\mathcal{L}(\theta)
=
\frac{1}{n}
\sum_{i=1}^n
\delta_{y_i}(z_i),
\end{equation}
where $\theta:=\{\psi,\pi,\sigma^2\}$, and $\delta_c(z)$ is given by \eqref{eq:delta-simplex}.

Unlike the fully free case in Section~\ref{sec:mle}, the class means cannot collapse: their locations and pairwise distances are fixed by the simplex construction. 
In our implementation, the simplex means $\{\mu_c\}$ are stored as a non-trainable buffer, the log-variance $\log \sigma^2$ and the prior logits are free parameters, and the forward pass returns the $\delta_c(z)$ values in \eqref{eq:delta-simplex} for each sample and class.

\paragraph{Why maximum likelihood.}
A central goal of this work is to train the Deep LDA head by maximum likelihood (equivalently, minimizing negative log-likelihood).
Prior ``Deep LDA'' style approaches typically optimize modified Fisher-type criteria rather than the LDA likelihood directly~\citep{DBLP:journals/corr/DorferKW15,DBLP:journals/pr/WuSH17,chen2021regularized}, or use discriminative objectives defined over Gaussian/Mahalanobis distances~\citep{pang2018maxmahalanobis}.
We make this distinction explicit because our contribution is precisely to obtain a well-posed likelihood-based formulation under simple geometric constraints.

\paragraph{Training} follows standard end-to-end maximum-likelihood optimization:
we fix the simplex means $\{\mu_c\}$, and jointly update the encoder parameters $\psi$ and head parameters
$(\pi,\sigma^2)$ by minimizing the negative log-likelihood (7) with Adam.
Unlike the batchwise re-estimation scheme in Section~2.3, there is no moment-matching: all learnable
parameters are updated by backpropagation. Our implementation is available at \url{https://github.com/zh3nis/simplex-lda}.

\paragraph{Results.}
On synthetic data generated from \eqref{eq:priors}--\eqref{eq:class_cond}, we apply early stopping, as the training accuracy reaches 99.2\% after only three epochs. The resulting Deep LDA model yields separated class clusters in the latent space and attains 99.4\% test accuracy; see Figure~\ref{fig:deep_lda_emb3}.\footnote{In an experiment replacing the log-likelihood with Fisher’s generalized eigenvalue objective in the simplex-constrained Deep LDA model produced similar embeddings and nearly identical train \& test accuracies.}

Having validated the approach on synthetic data, we now proceed to real-data evaluation.

\section{Experiments on Real Data}
\label{sec:real-data}

We now evaluate the proposed Deep LDA model of Section~\ref{sec:success-deep-lda} on image classification tasks.  
Across all experiments, the LDA head uses the same structure as described in Section~\ref{sec:success-deep-lda}. The encoder $f_\psi$ is trained jointly with $(\pi,\sigma^2)$ by maximizing the log-likelihood \eqref{eq:deep-lda-mle}. We compare this model to an identical encoder equipped with a conventional linear softmax layer trained via cross-entropy.  
To visualize the learned geometry, we additionally project the encoder output for a sample of trained embeddings to 2D via PCA.


\paragraph{Datasets.}
We evaluate on three standard image classification benchmarks.  
\textbf{Fashion-MNIST}~\citep{xiao2017fashion} contains 70\,000 grayscale images of size $28\times 28$ from 10 classes, split into 60\,000 training and 10\,000 test samples.  
\textbf{CIFAR-10} and \textbf{CIFAR-100}~\citep{krizhevsky2009learning} consist of 32{\(\times\)}32 color images with 10 and 100 classes respectively, each with 50\,000 training and 10\,000 test images. Following common practice, we apply standard data augmentation to
CIFAR-10/100 (random horizontal flips and random crops with padding)
and use only per-pixel normalization on Fashion-MNIST.

\paragraph{Encoder architecture.}
All experiments use the same convolutional encoder architecture, differing only in the number of input channels: one channel for Fashion-MNIST and three channels for CIFAR.  
The encoder consists of three convolutional blocks with channel widths $\{64,128,256\}$.  
Each block applies two $3\times 3$ $\mathrm{Conv}\!-\!\mathrm{BN}\!-\!\mathrm{ReLU}$ layers, with the first two blocks followed by $2\times 2$ max pooling.  
The final block is followed by global spatial aggregation via adaptive average pooling, producing a 256-dimensional vector.  
A linear layer projects this vector to the embedding dimension $d$:
\begin{multline*}
\mathrm{Conv}(C_{\mathrm{in}}\!\to\!64)
\to
\mathrm{Conv}(64\!\to\!64)
\to
\mathrm{Pool}
\to
\mathrm{Conv}(64\!\to\!128)
\to
\mathrm{Conv}(128\!\to\!128)\\
\to
\mathrm{Pool}
\to
\mathrm{Conv}(128\!\to\!256)
\to
\mathrm{Conv}(256\!\to\!256)
\to
\mathrm{AdaptiveAvgPool}
\to
\mathrm{Linear}(256\!\to\!d),
\end{multline*}
where $C_{\mathrm{in}}=1$ for Fashion-MNIST and $C_{\mathrm{in}}=3$ for CIFAR-10/100.
This encoder is used unchanged across all experiments, with the output dimension set to $d=C-1$ for Deep LDA and matched to the classifier input dimension for the softmax baseline.

\paragraph{Heads and training.}
For each dataset we compare two models built on the same encoder.  
The first uses a standard linear softmax layer $\mathbb{R}^d \!\to\! \mathbb{R}^C$ trained with cross-entropy.  
The second replaces the softmax with our structurally constrained LDA head trained with MLE and Fisher's objective.  
Both models are optimized with Adam, batch size 256, for 100~epochs on all three datasets.

\paragraph{Results.}
Table~\ref{tab:image-results} shows test accuracies averaged over three seeds.
On Fashion-MNIST, LDA matches softmax almost exactly.
On CIFAR-10, LDA slightly outperforms softmax.
On CIFAR-100, the two heads achieve essentially identical accuracy.
Overall, our structured LDA is competitive with softmax across datasets while producing substantially cleaner and more structured latent representations.

\setlength{\tabcolsep}{5pt}
\begin{table}[htbp]
  \centering
  \caption{Image classification: Test accuracies (mean $\pm$ std over 3 runs).}
  \label{tab:image-results}
  \begin{tabular}{llccc}
    \toprule
    Head & Objective & Fashion-MNIST & CIFAR-10 & CIFAR-100  \\
    \midrule
    Softmax & Cross-entropy loss & $93.61\pm.38$ & $89.07\pm.20$ & $64.67\pm.30$ \\
    Simplex LDA & Log-likelihood & $93.40\pm.32$ & $90.41\pm.04$ & $64.61\pm.12$ \\
    Simplex LDA & Fisher's objective & $92.94\pm.07$ & $90.97\pm.37$ & $64.71\pm.15$\\
    \bottomrule
  \end{tabular}
\end{table}



\begin{figure}[htbp]
  \centering

  \begin{subfigure}[b]{0.49\linewidth}
    \centering
    \includegraphics[width=\linewidth]{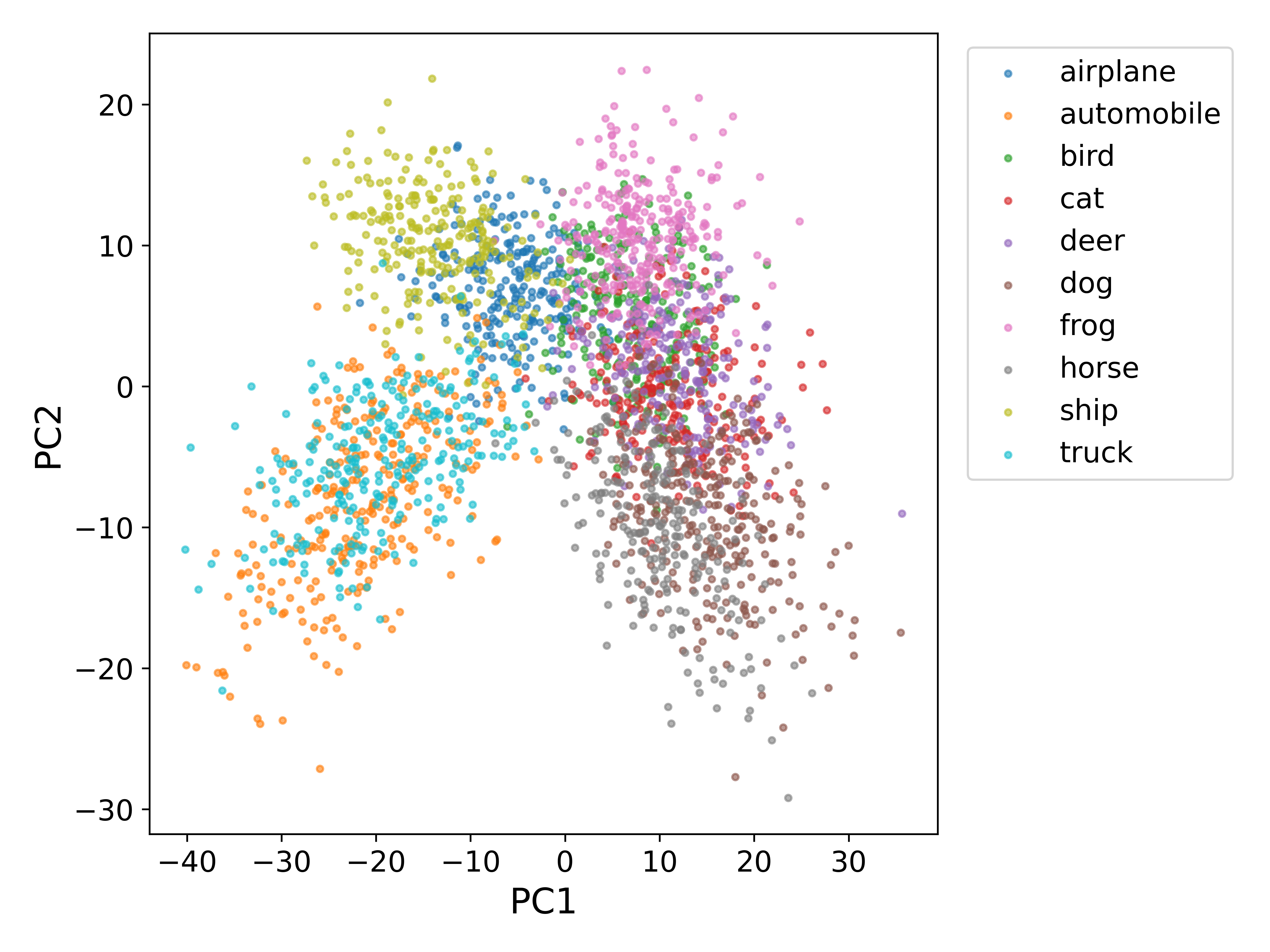}
    \caption{Softmax.}
  \end{subfigure}
  \hfill
  \begin{subfigure}[b]{0.49\linewidth}
    \centering
    \includegraphics[width=\linewidth]{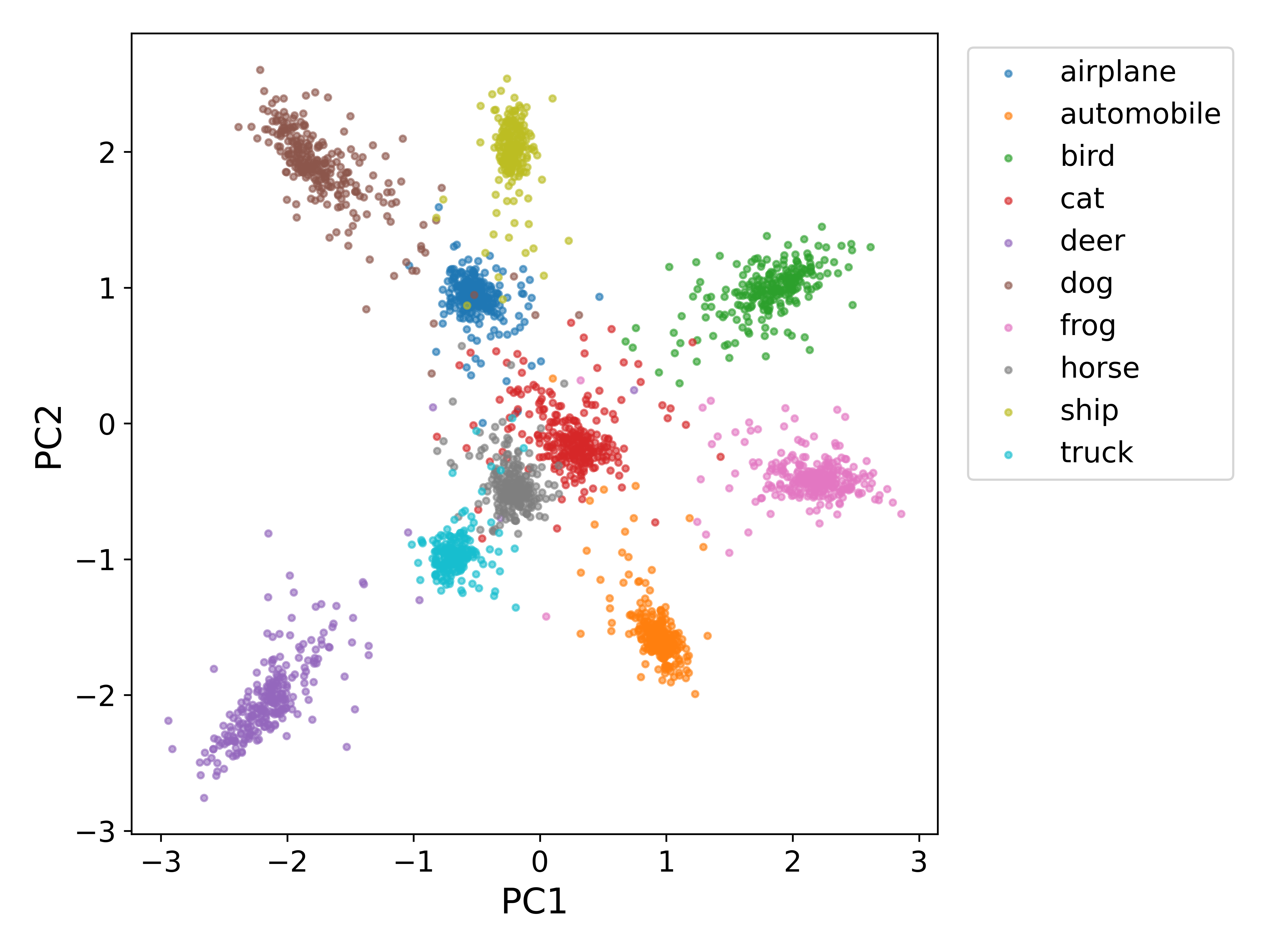}
    \caption{Simplex LDA.}
  \end{subfigure}
  \caption{PCA projections of CIFAR-10 embeddings learned with two classification heads.}
  \label{fig:img-embeddings-2d}
\end{figure}

\paragraph{Embedding Geometry.}
Figure~\ref{fig:img-embeddings-2d} compares PCA projections of the learned
embeddings for the softmax and LDA heads. Under the softmax head, the
projected embeddings form clusters that partially
overlap. In contrast, the LDA
head yields very tight class clusters that remain separated even after projection on 2D. This illustrates that LDA imposes a strong geometric prior:
maximizing between-class separation while shrinking within-class
variance, leading to a much more structured and discriminative
representation.

\begin{wrapfigure}{r}{.45\textwidth}
\centering
  \includegraphics[width=.4\textwidth]{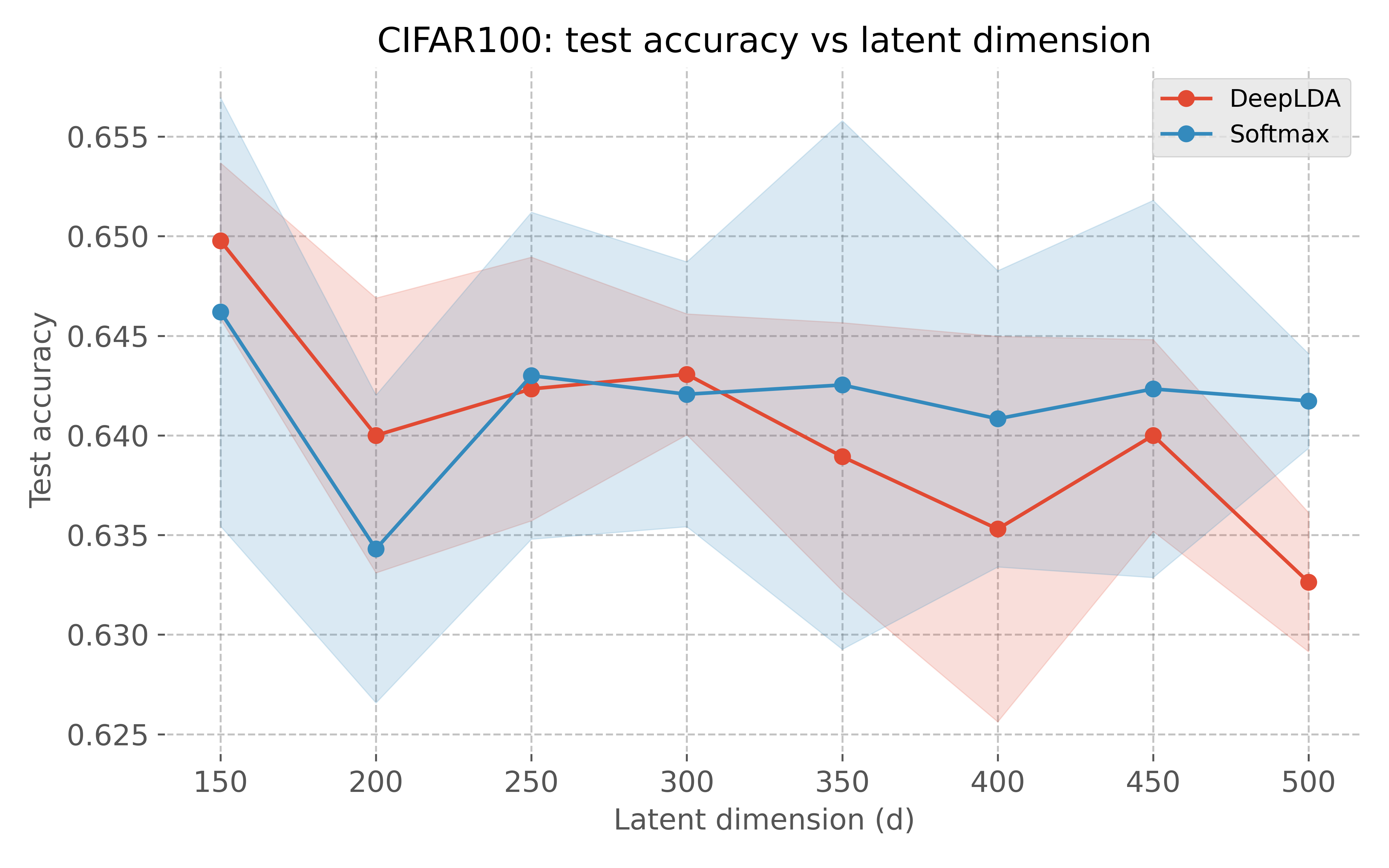}
  \caption{CIFAR-100 test accuracy as a function of latent dimension $d$ for the Softmax and LDA heads (mean $\pm 2$ standard deviations over three runs).}
  \label{fig:cifar100-dim-sweep}
\end{wrapfigure}

\paragraph{Effect of latent dimension.}
In the main experiments we set the latent dimension to $d = C-1$, which is the smallest dimension that can accommodate a $(C-1)$-dimensional simplex.
To check the effect of the latent dimension, we ran an additional experiment on CIFAR-100 by varying $d \in \{150, 200, \dots, 500\}$. Figure~\ref{fig:cifar100-dim-sweep}
reports test accuracy as a function of $d$.
Softmax accuracy is almost insensitive to the choice of $d$, whereas Deep LDA exhibits a mild decreasing trend as $d$ increases: enlarging the latent space offers no benefit and can slightly hurt performance.
Overall, both heads remain within a narrow accuracy band for all tested dimensions, and Deep LDA is competitive with softmax for moderate $d$.
This suggests that the generative head actually prefers compact latent spaces and supports our use of the minimal dimension $d = C-1$ as a parsimonious and empirically favorable setting.

\paragraph{Covariance structure ablation.}
Our default head uses a spherical shared covariance, $\Sigma=\sigma^2 I$, to keep the head lightweight and comparable in parameter count to a linear softmax layer at the same embedding dimension.
To test whether additional covariance flexibility helps, we train simplex-based Deep LDA with spherical, diagonal, and full covariances on FashionMNIST, CIFAR-10, and CIFAR-100 (3 runs per setting).
Table~\ref{tab:cov-ablation} reports mean $\pm$ std test accuracy.
Neither diagonal nor full covariance improves performance; in fact, full covariance significantly degrades accuracy on CIFAR-100.
This suggests that the encoder can adapt to a spherical covariance structure, while additional covariance degrees of freedom may introduce optimization/estimation difficulties.

\begin{table}[htbp]
\centering
\setlength{\tabcolsep}{6pt}
\caption{Ablation of the shared covariance parameterization in simplex-based Deep LDA (mean $\pm$ std over 3 runs).}
\begin{tabular}{lccc}
\toprule
Covariance & FashionMNIST & CIFAR-10 & CIFAR-100 \\
\midrule
Spherical & $93.38 \pm 0.06$ & $90.49 \pm 0.26$ & $65.20 \pm 0.22$ \\
Diagonal  & $93.32 \pm 0.07$ & $90.31 \pm 0.26$ & $65.02 \pm 0.09$ \\
Full      & $93.31 \pm 0.11$ & $89.91 \pm 0.31$ & $59.64 \pm 0.71$ \\
\bottomrule
\end{tabular}
\label{tab:cov-ablation}
\end{table}

\paragraph{Stronger off-the-shelf backbone (ResNet-50).}
We additionally evaluate our approach with a stronger, widely used backbone.
Specifically, we fine-tune an ImageNet-pretrained ResNet-50 encoder and compare (i) our Simplex LDA likelihood-trained head against (ii) a standard Softmax head trained with cross-entropy.
To ensure a controlled comparison, we use the \emph{same} encoder, the \emph{same} embedding dimension for both heads, setting $D=C-1$.
Table~\ref{tab:resnet50} reports results over 3 runs (mean $\pm$ std).
Overall, with a substantially stronger backbone, the simplex-constrained likelihood-trained head remains competitive with the conventional softmax baseline across datasets.
\begin{table}[htbp]
\centering
\setlength{\tabcolsep}{6pt}
\caption{Fine-tuning a pretrained ResNet-50 backbone (mean $\pm$ std over 3 runs).}
\begin{tabular}{lccc}
\toprule
Head & FashionMNIST & CIFAR-10 & CIFAR-100 \\
\midrule
Softmax     & $93.99 \pm 0.17$ & $96.30 \pm 0.14$ & $80.85 \pm 0.19$ \\
Simplex LDA & $94.22 \pm 0.08$ & $96.35 \pm 0.19$ & $80.45 \pm 0.22$ \\
\bottomrule
\end{tabular}
\label{tab:resnet50}
\end{table}

\paragraph{Beyond images: text classification with a pretrained Transformer.}
To demonstrate that the proposed head generalizes beyond image benchmarks, we fine-tune a
pretrained Transformer (BERT-base-uncased \cite{devlin2019bert}) for text classification on AG News
(4 classes; 120k train / 7.6k test)\footnote{\url{http://groups.di.unipi.it/~gulli/AG_corpus_of_news_articles.html}} and CLINC150 (151 classes; 15k train / 5.5k test) \cite{larson-etal-2019-evaluation}.
We compare (i) a standard linear Softmax head to (ii) our SimplexLDA head, using an
architecture-matched setup with an embedding projection to $d=C-1$.
Both models are fine-tuned with AdamW (learning rate $2\times 10^{-5}$, batch size 32) for
3 epochs, and results are averaged over 3 random seeds.
Table~\ref{tab:text} reports test accuracy (mean $\pm$ std).
Overall, SimplexLDA remains competitive with Softmax on AG News and improves performance on CLINC150,
supporting the generality and scalability of the proposed likelihood-trained head.

\begin{table}[htbp]
\centering
\caption{Text classification: test accuracy (mean $\pm$ std over 3 runs).}
\label{tab:text}
\begin{tabular}{lcc}
\toprule
Head & AG News & CLINC150 \\
\midrule
Softmax    & $94.40 \pm 0.07$ & $87.11 \pm 0.56$ \\
SimplexLDA & $94.37 \pm 0.37$ & $88.12 \pm 0.34$ \\
\bottomrule
\end{tabular}
\end{table}

\section{Related Work}
\label{sec:related}

Classical Linear Discriminant Analysis dates back to \citet{fisher1936use} and \citet{rao1948}, who modeled each class as a Gaussian with a shared covariance matrix, yielding linear decision boundaries. Adapting LDA to deep networks has inspired several research directions, most of which modify the classical objective rather than preserve its generative form.

\paragraph{Deep Fisher-style objectives.}
DeepLDA~\citep{DBLP:journals/corr/DorferKW15} optimizes Fisher-style generalized eigenvalue criterion, with related extensions for person re-identification~\citep{DBLP:journals/pr/WuSH17} and
regularized scatter objectives~\citep{chen2021regularized}. In its naive form, maximizing the
mean eigenvalue of the LDA problem causes the network to favor trivial solutions that
magnify a single dominant eigenvalue while neglecting poorly separated classes; this behavior
was explicitly observed by \citet{DBLP:journals/corr/DorferKW15}, who therefore replace the original objective with a
heuristic that maximizes only the $k$ smallest eigenvalues (their Eqs.~(8)--(9)). As a result,
these methods promote between/within-class separation but no longer correspond to maximum
likelihood under a generative LDA model, in contrast to our constrained Deep LDA formulation.

\paragraph{Fixed classifiers and simplex-structured heads.}
A complementary line of work studies \emph{fixing} the final classifier layer to impose a prescribed geometry.
\citet{hoffer2018fix} argue that the last linear layer can be fixed (up to scale) with little loss in accuracy.
Building on this idea, \citet{pernici2022regular} fix class vectors to the vertices of regular polytopes (including the simplex) to obtain maximally separated and stationary embeddings. Motivated by the neural collapse phenomenon \cite{papyan2020prevalence}, several works explicitly use a regular simplex (simplex equiangular tight frame) structure for the classifier and/or class centers, either by analyzing the induced optimization landscape~\citep{zhu2021geometric}, by demonstrating that fixing a simplex ETF classifier can induce neural collapse even under class imbalance~\citep{yang2022inducing}, or by proposing to optimize \emph{towards} the nearest simplex ETF rather than hard-fixing one for improved stability and convergence~\citep{markou2024guiding}. Importantly, in these works the simplex (or polytope) structure is primarily leveraged to improve \emph{discriminative} training with losses such as cross-entropy or squared error.

\paragraph{Our contribution.}
Prior deep LDA variants either modify Fisher’s criterion or use Gaussian distances in a discriminative loss, and fixed/simplex classifier works use simplex structure mainly as an inductive bias for discriminative training. In contrast, we retain the generative LDA model and train it by maximum likelihood. In our setting, the regular simplex plays a different role: it acts as a natural constraint that makes end-to-end likelihood training of Deep LDA well-posed, preventing the degeneracies that arise under an unconstrained model (Section~\ref{sec:mle}). Together with a spherical shared covariance, these constraints yield an interpretable, well-behaved generative classifier whose MLE training produces separated class clusters without collapse.

\section{Conclusion}
We revisited Deep Linear Discriminant Analysis from a likelihood perspective and showed that the unconstrained deep model is ill-posed under maximum-likelihood training: the encoder collapses embeddings, some of the class means drift toward each other, and the shared covariance shrinks toward singularity, yielding high likelihood but poor discrimination. A simple geometric constraint---fixing the class means to a regular simplex and enforcing a spherical shared covariance---removes the pathological tendency of the class clusters to collapse or heavily overlap, and improves the classification performance. On synthetic data and image benchmarks, the resulting Deep LDA classifier yields clean, interpretable latent spaces with well-separated clusters and accuracy competitive with softmax, illustrating that mild geometric structure can restore the viability of likelihood-based training for deep discriminant analysis. We additionally verified that the head transfers beyond images by fine-tuning a pretrained Transformer on text classification (AG News, CLINC150), where it remains competitive with Softmax.

\section*{Acknowledgements}
This work was supported by the Science Committee of the Ministry of Science and Higher Education of the Republic of Kazakhstan (Grant No. AP26198325).

\paragraph{Reproducibility.} Code to reproduce our experiments is available at \url{https://github.com/zh3nis/simplex-lda}.

\paragraph{AI Assistance.} We used an AI-based tool for editorial polishing of the text; the authors take full responsibility for the content.

\bibliography{ref}

\end{document}